\documentclass[runningheads]{llncs}
\usepackage{graphicx}
\usepackage{tikz}
\usepackage{comment}
\usepackage{amsmath,amssymb} % define this before the line numbering.
\usepackage{color}
\usepackage{pifont}
\usepackage{bbm}
\usepackage{bbding}
\usepackage{multirow}
\usepackage{pifont}
\usepackage{subfigure}  
\usepackage{booktabs}
\usepackage{paralist} % compactenum
\usepackage[accsupp]{axessibility}  % Improves PDF readability for those with disabilities.

\makeatletter
\newcommand{\printfnsymbol}[1]{%
        \textsuperscript{\@fnsymbol{#1}}%
}
\makeatother

\begin{document}
% \renewcommand\thelinenumber{\color[rgb]{0.2,0.5,0.8}\normalfont\sffamily\scriptsize\arabic{linenumber}\color[rgb]{0,0,0}}
% \renewcommand\makeLineNumber {\hss\thelinenumber\ \hspace{6mm} \rlap{\hskip\textwidth\ \hspace{6.5mm}\thelinenumber}}
% \linenumbers
\pagestyle{headings}
\mainmatter

\title{SimCC: a Simple Coordinate Classification Perspective for Human Pose Estimation} % Replace with your title

% CAMERA READY SUBMISSION
% \begin{comment}
\titlerunning{Simple Coordinate Classification Perspective for Human Pose Estimation}
% If the paper title is too long for the running head, you can set
% an abbreviated paper title here

\author{preprint}
\author{Yanjie Li$^1$ \quad Sen Yang$^2$ \quad
Peidong Liu$^1$ \quad
Shoukui Zhang$^3$ \quad
Yunxiao Wang$^1$ \quad
Zhicheng Wang$^{4}$\thanks{Corresponding authors.} \quad
Wankou Yang$^2$ \quad
Shu-Tao Xia$^{1,5}$\printfnsymbol{1}
}
\authorrunning{Y. Li, S. Yang, P. Liu et al.}
% First names are abbreviated in the running head.
% If there are more than two authors, 'et al.' is used.
%
\institute{$^1$Tsinghua University \quad $^2$Southeast University \quad $^3$Meituan Inc. \quad $^4$Nreal \\ $^5$PCL Research Center of Networks and Communications, Peng Cheng Laboratory \\
}
% \end{comment}
%******************
\maketitle

\begin{abstract}
The 2D heatmap-based approaches have dominated Human Pose Estimation (HPE) for years due to high performance. However, the long-standing quantization error problem in the 2D heatmap-based methods leads to several well-known drawbacks: 1) The performance for the low-resolution inputs is limited; 2) To improve the feature map resolution for higher localization precision, multiple costly upsampling layers are required; 3) Extra post-processing is adopted to reduce the quantization error. To address these issues, we aim to explore a brand new scheme, called \textit{SimCC}, which reformulates HPE as two classification tasks for horizontal and vertical coordinates. The proposed SimCC uniformly divides each pixel into several bins, thus achieving \emph{sub-pixel} localization precision and low quantization error. Benefiting from that, SimCC can omit additional refinement post-processing and exclude upsampling layers under certain settings, resulting in a more simple and effective pipeline for HPE. Extensive experiments conducted over COCO, CrowdPose, and MPII datasets show that SimCC outperforms heatmap-based counterparts, especially in low-resolution settings by a large margin.

\keywords{Human Poes Estimation, 2D Heatmap, Coordinate Classification}
\end{abstract}

\section{Introduction}
2D Human Pose Estimation~(HPE) aims to localize body joints from a single image, where 2D heatmap-based methods~\cite{heatmap:cai2020learning,heatmap:cao2019openpose,heatmap:chen2018cascaded,heatmap:cheng2020higherhrnet,heatmap:darkpose,heatmap:li2019rethinking,heatmap:li2021tokenpose,heatmap:luo2020rethinking,heatmap:newell2016stacked,heatmap:sun2019deep,heatmap:xiao2018simple,heatmap:yang2020transpose} has become the \emph{de facto} standard in recent years. The 2D heatmap is generated as a 2-dimensional Gaussian distribution centering at the ground-truth joint position, which inhibits the cases of false positive and smooths the training process by assigning a probability value to each position.

\begin{figure}[t]
\centering
\includegraphics[width=\linewidth]{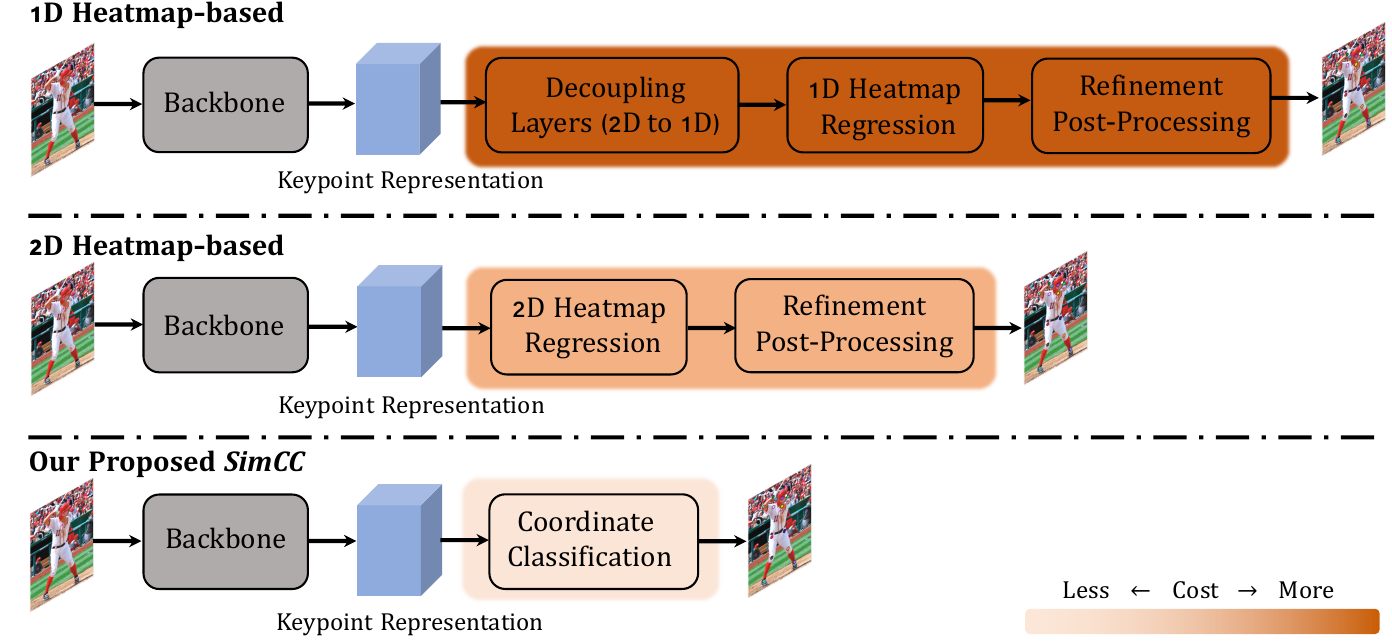} 
\vspace{-5mm}
\caption{\textbf{Comparisons between the proposed SimCC and 2D/1D heatmap-based pipelines}. The 2D heatmap-based scheme includes: 1) a backbone to extract keypoint representations; 2) a regression head to generate the 2D heatmap, which may consist of multiple time-consuming upsampling layers; 3) extra post-processing to refine the predictions, such as empirical shift and DARK~\cite{heatmap:darkpose}. The 1D heatmap regression~\cite{yin2020attentive} is introduced for facial landmark. Compared to the 2D heatmap-based scheme, The 1D heatmap regression~\cite{yin2020attentive} brings additional learnable decoupling layers consisting of multiple CNN layers and a co-attention module, to transform the 2D features to 1D heatmaps. Different from these heatmap-based schemes, the proposed SimCC is much simpler, which only needs two classifier heads for coordinate classification and excludes the costly refinement post-processing and extra upsampling layers.}
\label{fig:comparison}
\end{figure}

Despite its success, heatmap-based methods suffer seriously from the long-standing quantization error problem, which is caused by mapping the continuous coordinate values into discretized 2D downscaled heatmaps. The substantial quantization error further brings about several well-known shortcomings: 1) Costly upsampling layers (\emph{e.g.}, deconvolution layers~\cite{heatmap:xiao2018simple}) are used to increase the feature map resolution to alleviate the quantization error;
2) Extra post-processing (\emph{e.g.}, empirical shift, DARK~\cite{heatmap:darkpose}) is introduced to refine the predictions;
3) The performances are far from satisfactory for low-resolution inputs due to the serious quantization error. Considering obtaining high-resolution 2D heatmap brings heavy computation cost, a natural way to decrease the quantization error is to firstly decouple the 2D heatmap into 1D heatmap and then increase the resolution, which has been explored by Yin et al.~\cite{yin2020attentive} for facial landmark area. However, to realize that goal, Yin et al.~\cite{yin2020attentive} introduces additional decoupling layers (\emph{i.e.}, a co-attention module and multiple convolution layers) and costly deconvolution modules, resulting in an even more complicated 
pipeline than 2D heatmap-based methods, as shown in Fig.~\ref{fig:comparison}.

Therefore, in this work, we try to explore a brand-new scheme against heatmap-based methods for HPE. Specifically, we propose a simple yet effective coordinate classification pipeline, namely \emph{SimCC}, which regards HPE as two classification tasks for horizontal and vertical coordinates. SimCC firstly employs a Convolutional Neural Network (CNN) or Transformer-based backbone to extract keypoint representations. Given the obtained keypoint representations, SimCC then performs coordinate classification for vertical and horizontal coordinates independently to yield the final predictions. To reduce the quantization error, SimCC uniformly divides each pixel into several bins, which achieves \emph{sub-pixel} localization precision. Note that different from heatmap-based approaches which may introduce multiple deconvolution layers, SimCC only needs two lightweight classifier heads (\emph{i.e.} only one linear layer for each head).

Fig.~\ref{fig:comparison} shows the comparisons between our proposed SimCC and 1D/2D heatmap-based approaches. Compared to the dominant 2D heatmap-based scheme, SimCC has three benefits: 1) It reduces quantization error by uniformly dividing each pixel into several bins; 2) SimCC omits upsampling layers under certain settings~\cite{heatmap:xiao2018simple} and excludes the costly refinement post-processing, which is more friendly to real-world applications; 3) SimCC shows impressing performance even with low input sizes. Our contributions are summarized as follows:

\begin{compactenum}[--]
    \item We propose a coordinate classification pipeline for human pose estimation called SimCC, reformulating the problem as two classification tasks for horizontal and vertical coordinates. SimCC serves as a general scheme and can be easily applied to existing CNN-based or Transformer-based HPE models.
    \item SimCC achieves high efficiency by omitting the extra time-consuming upsampling and post-processing in heatmap-based methods. In particular, applying SimCC reduces over 55\% GFLOPs of SimBa-Res50~\cite{heatmap:xiao2018simple} and achieves higher model performance than heatmap-based counterpart.
        \item Comprehensive experiments over COCO, CrowdPose, and MPII datasets are conducted to verify the effectiveness of the proposed SimCC with different backbones and multiple input sizes.
\end{compactenum}

It's our belief that the predominant 2D heatmap-based methods may not be the final solution for HPE due to its high computation cost, complicated post-processing and poor performance under low input resolutions. We hope that the exploration of SimCC could provide a new perspective for the potential research work and practical deployment for HPE.

%%%%%%%%%%%%%%%%%%%%%%%%%%%%%%% related work
\section{Related Work}
%%%%%%%%%%%%%%%%%%%%%%%%%%%%%%%  regression-based hpe
\vspace{.1cm}
\noindent\textbf{Regression-based HPE.} 
Regression-based methods~\cite{toshev2014deeppose,carreira2016human,tian2019directpose,sun2018integral,sun2017compositional,nie2019single} are explored more often in the early stage of 2D human pose estimation. Different from relying on 2D grid-like heatmap, this line of work directly regresses the keypoint coordinates in a computationally friendly framework. However, only a handful of existing methods adopt this scheme due to the unsatisfactory performance. Very recently, Li \emph{et al}.~\cite{rle} introduce the residual log-likelihood (RLE), which utilizes the normalizing flows~\cite{flow} to capture the underlying output distribution and makes regression-based methods match the accuracy of state-of-the-art heatmap-based
methods. The core idea of RLE is to construct an adaptive loss based on the normalizing flows and therefore complementary to our work.

%%%%%%%%%%%%%%%%%%%%%%%%%%%%%%% heatmap-based hpe
\vspace{.1cm}
\noindent\textbf{2D heatmap-based HPE.} 
Another line of work~\cite{heatmap:cai2020learning,heatmap:cao2019openpose,heatmap:chen2018cascaded,heatmap:cheng2020higherhrnet,heatmap:li2019rethinking,heatmap:li2021tokenpose,heatmap:luo2020rethinking,heatmap:newell2016stacked,heatmap:sun2019deep,heatmap:xiao2018simple,heatmap:yang2020transpose,heatmap:darkpose,udp} adopts two-dimensional Gaussian distribution (\emph{i.e.}, \emph{heatmap}) to represent joint coordinate. Each position on the heatmap is assigned with a probability to be the ground truth point. As one of the earliest uses of heatmap, Tompson \emph{et al.}~\cite{heatmap:tompson2014heatmap} propose a hybrid architecture consisting of a deep Convolutional Network and a Markov Random Field. Newell \emph{et al.}~\cite{heatmap:newell2016stacked} introduce hourglass-style architecture into HPE. Papandreou \emph{et al.}~\cite{mppe} propose to aggregate the heatmap and offset prediction to improve the localization precision. Xiao \emph{et al.}~\cite{heatmap:xiao2018simple} propose a simple baseline that utilizes three deconvolutional layers following a backbone network to obtain the final predicted heatmap. Instead, Sun \emph{et al.}~\cite{heatmap:sun2019deep} propose a novel network to maintain high-resolution representations through the whole process, achieving significant improvement. Owing to the involvement of spatial uncertainty, this kind of learning schema has the tolerance of mistakes of jitter. As a result, heatmap-based methods keep stable state-of-the-art performance for years. However, quantization error remains a significant problem of the heatmap-based methods, especially in low input resolutions.

%%%%%%%%%%%%%%%%%%%%%%%%%%%%%%% heatmap-based hpe
\vspace{.1cm}
\noindent\textbf{quantization error.} To address the large quantization error caused by the discretized 2D downscaled heatmaps,  Zhang \emph{et al.}~\cite{heatmap:darkpose} propose to comprehensively account for the distribution information of heatmap activation by adopting Taylor-expansion based distribution approximation as post-processing. Another attempt to alleviate the quantization error is explored by Yin \emph{et al}.~\cite{yin2020attentive} for facial landmark detection, which adopts learnable decoupling layers to transform 2D heatmap to 1D heatmap and then uses additional deconvolution layers to increase the 1D heatmap resolution.

\vspace{.1cm}
\noindent\textbf{1D heatmap regression in facial landmark.} Outside the realm of human pose estimation, 1D heatmap-based methods~\cite{yin2020attentive,gv} have been explored for facial landmark detection. Among those, Yin \emph{et al}.~\cite{yin2020attentive} propose an attentive 1D heatmap regression method, aiming to alleviate the quantization error by using deconvolution layers to raise up the decoupled 1D heatmaps resolution. To capture the joint distribution information between the decoupled 1D heatmaps, a co-attention module is introduced in the 1D heatmap regression~\cite{yin2020attentive}. 

\vspace{.1cm}
\noindent\textbf{Coordinate classification.} Concurrent to our work,  Chen \emph{et al}.~\cite{chen2021pix2seq} propose Pix2Seq to casts object detection as a language modeling task, where an object is described as sequences of five discrete tokens for further classification. In Pix2Seq, the Transformer decoder architecture is essential to ``read out'' each object (yield the predictions). By contrast, our proposed SimCC aims to explore a new path against heatmap-based methods for human pose estimation, which can be easily combined with CNN or Transformer-based HPE methods and does not rely on an additional Transformer decoder to generate the prediction.

\begin{figure*}[t]
	\centering
\includegraphics[width=\linewidth]{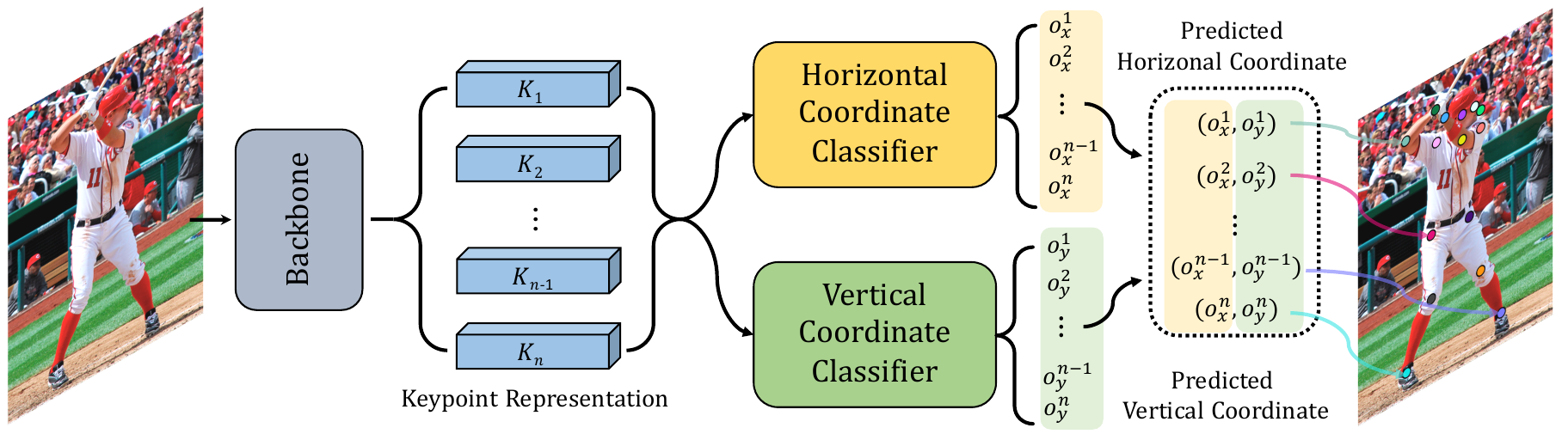} 
	\caption{\textbf{The proposed SimCC pipeline for HPE}. SimCC firstly extracts $n$ keypoint representations via a backbone which can be either CNN-based or Transformer-based networks. For the CNN-based backbone, we simply flatten the obtained keypoint representations from \footnotesize{$(n, H', W')$} to \footnotesize{$(n, H'\times W')$} for the subsequent classification. Based on the $n$ keypoint representations, SimCC then performs coordinate classification for horizontal and vertical axes independently to yield the final predictions. Specifically, given $i$-th keypoint representation as input, the horizontal and vertical coordinate classifiers (\emph{i.e.}, only one linear layer for each classifier) generate the $i$-th keypoint predictions $o_x^i$ and $o_y^i$, respectively. Note that SimCC uniformly divides each pixel into multiple bins thus the quantization error is reduced and sub-pixel localization precision is achieved.}
	\label{fig:pipeline}
\end{figure*}

\section{SimCC: Reformulating HPE from Classification Perspective}
\label{SimCC}

The key idea of SimCC is to regards human pose estimation as two classification tasks for vertical and horizontal coordinates and to reduce quantization error by dividing each pixel to multiple bins. Fig.~\ref{fig:pipeline} shows the schematic illustration of SimCC composed of a backbone network and two classifier heads. We will describe each components in this section in details.

\noindent\textbf{Backbone.} Given an input image of size $H\times W\times3$, SimCC employs either CNN-based or Transformer-based network (\emph{e.g.}, HRNet~\cite{heatmap:sun2019deep}, TokenPose~\cite{heatmap:li2021tokenpose}) as the backbone to extract $n$ keypoint representations for $n$ corresponding keypoints.

\noindent\textbf{Head.} As shown in Fig.~\ref{fig:pipeline}, horizontal and vertical classifiers (\emph{i.e.}, only one linear layer for each classifier) are appended after the backbone to perform coordinate classification, respectively. For the CNN-based backbone, we simply flatten the outputted keypoint representations from $(n, H', W')$ to $(n, H'\times W')$ for classification. Compared to heatmap-based approach~\cite{heatmap:xiao2018simple} which uses multiple costly deconvolution layers as head, SimCC head is much more lightweight and simple.

\noindent\textbf{Coordinate classification.} To achieve classification, we propose to uniformly discretize each continuous coordinate value into an integer as class label for model training: $c_x\in[1,N_x], c_y\in[1,N_y]$, where $N_x=W\cdot k$ and $N_y=H\cdot k$ represent the number of bins for horizontal and vertical axes, respectively. $k$ is the splitting factor and set as $\geq 1$ to reduce quantization error, resulting in \emph{sub-pixel} localization precision. To yield the final prediction, SimCC performs vertical and horizontal coordinate classification independently based on the $n$ keypoint representations learnt by the backbone. Concretely, given $i$-th keypoint representation as input, the $i$-th keypoint predictions $o_x^i$ and $o_y^i$ are generated by the horizontal and vertical coordinate classifiers, respectively. In addition, Kullback–Leibler divergence is used as loss function for training.

\noindent\textbf{Label smoothing.} In traditional classification tasks, label smoothing~\cite{labelSmoothing} is widely utilized to enhance model performance. Hence, we adopt it for SimCC, which is called \emph{equal label smoothing} in this paper. However, equal label smoothing punishes the false labels indiscriminately, which has ignored the spatial relevance of adjacent labels for the task of human pose estimation. A more reasonable solution is supposed to encourage the model to work in this way: the closer the output category is to the groundtruth, the better. To address this issue, we also explore to use \emph{Laplace} or \emph{Gaussian label smoothing}, resulting in smoothed labels following corresponding distribution. Unless noted otherwise, SimCC is used as the abbreviation for the variant with equal label smoothing .

\begin{figure}[t]
\centering
\subfigure[64$\times$64 input size.]{
\includegraphics[width=2.2in]{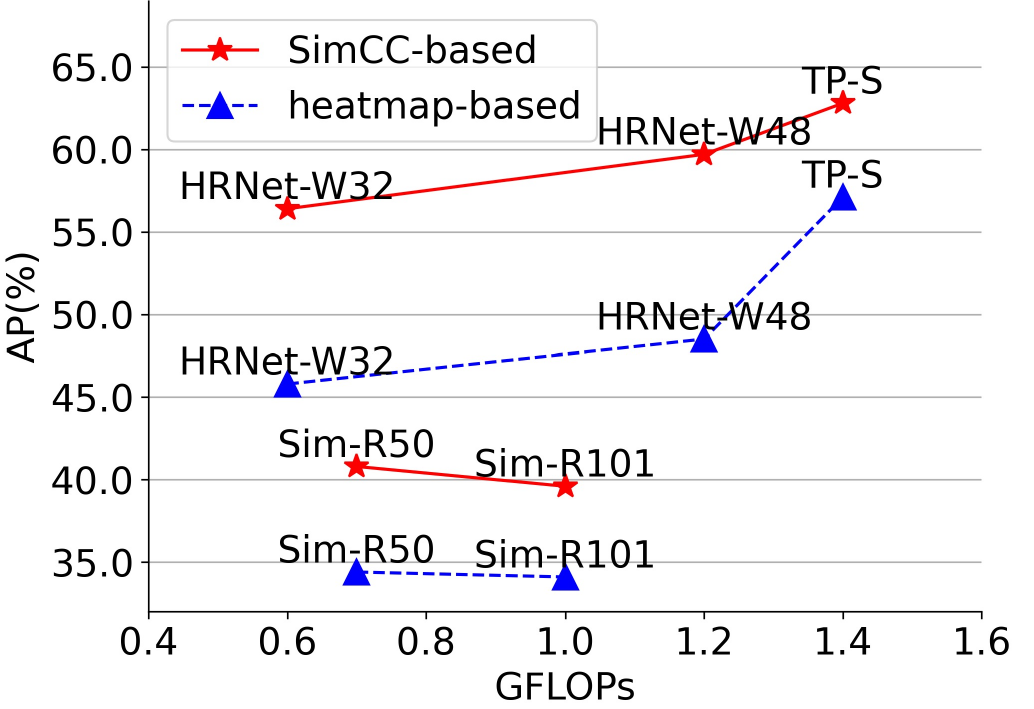}
}
\quad
\subfigure[128$\times$128 input size.]{
\includegraphics[width=2.2in]{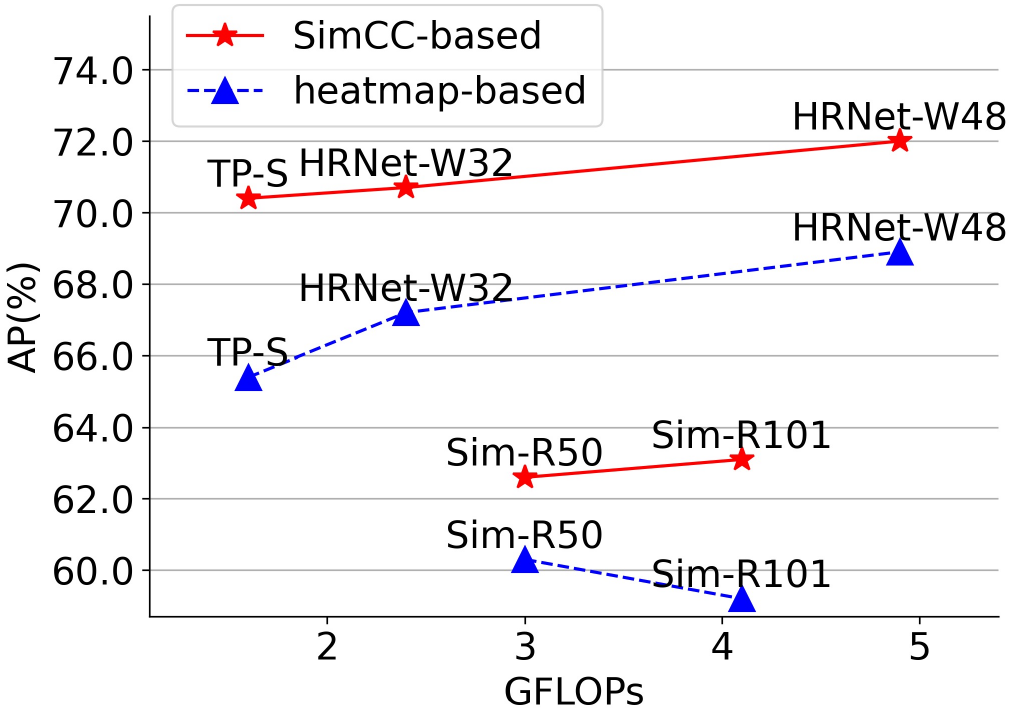}
}
\subfigure[256$\times$192 input size.]{
\includegraphics[width=2.2in]{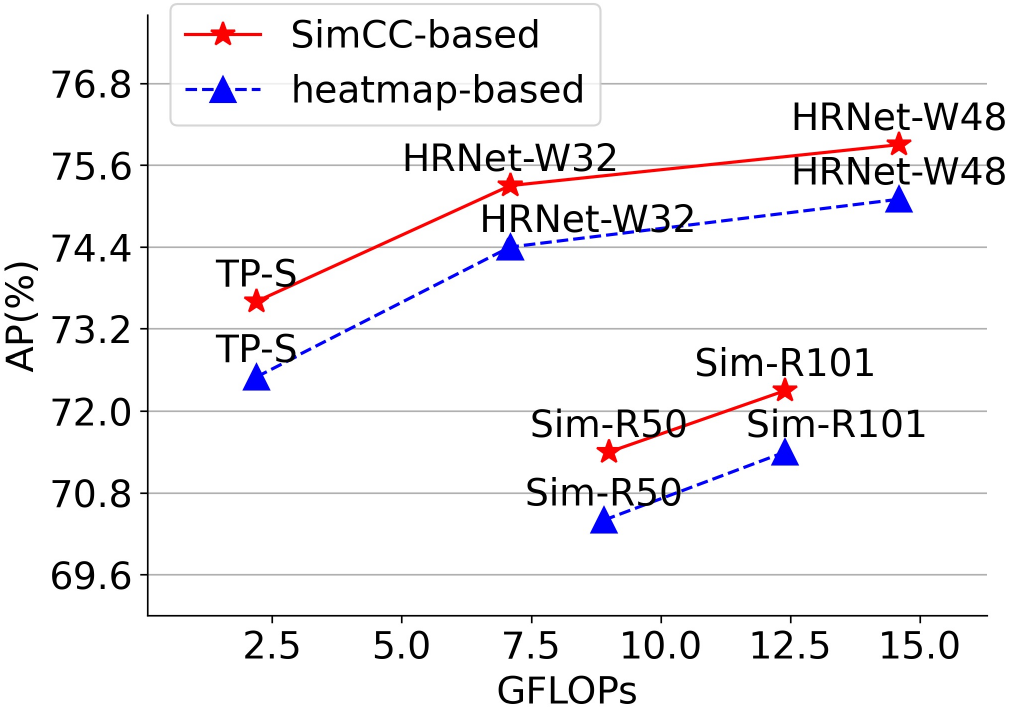}
}
\quad
\subfigure[384$\times$288 input size.]{
\includegraphics[width=2.2in]{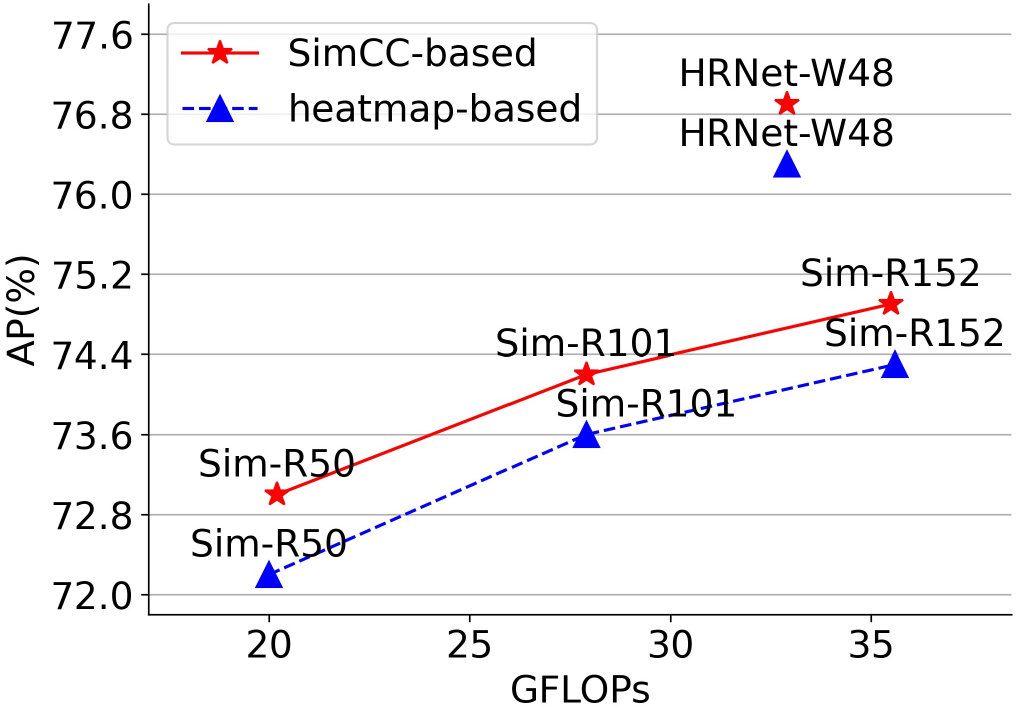}
}
\caption{\textbf{Comparisons with 2D heatmap-based approaches on COCO2017 val set across various input sizes}. `TP-S' and `Sim' represent TokenPose-S~\cite{heatmap:li2021tokenpose} and SimpleBaseline~\cite{heatmap:xiao2018simple}, respectively. Both CNN-based (\emph{i.e.}, SimpleBaseline~\cite{heatmap:xiao2018simple} and HRNet~\cite{heatmap:sun2019deep}) and Transformer-based (\emph{i.e.}, TokenPose~\cite{heatmap:li2021tokenpose}) HPE models are chosen to verify the effectiveness of the proposed SimCC. SimCC shows clear gains compared to 2D heatmap-based counterparts across various input sizes, especially in low resolution.}
\label{fig:robustness}
\end{figure}

\subsection{Comparisons to 2D heatmap-based approaches} 
In this part, we give a comprehensive investigation on the superiority of using SimCC scheme compared to the 2D heatmap-based approaches.

\noindent
\textbf{Quantization error.} Due to the computational cost of obtaining or maintaining high-resolution two-dimensional structure, 2D heatmap-based methods tend to output feature maps with $\lambda\times$ downscaled input resolution, which significantly enlarges the quantization error. On the contrary, SimCC uniformly divides each pixel into $k$ ($\geq1$)  bins during
discretization, which reduces the quantization error and obtains sub-pixel localization precision.

\noindent
\textbf{Refinement post-processing.} Heatmap-based approaches rely heavily on extra post-processing (\emph{e.g.}, empirical shift and DARK~\cite{heatmap:darkpose}) to reduce the quantization error. As shown in Table~\ref{tab:val}, the performance of heatmap-based methods drops significantly if without using post-processing for refinement. However, these post-processing strategies are usually computationally expensive and thus unfriendly to real-world applications. For example, DARK~\cite{heatmap:darkpose} uses Taylor-expansion and higher derivative needs to be calculated based on the obtained 2D heatmap. By contrast, the proposed SimCC omits refinement post-processing due to its sub-pixel localization precision, resulting in a simpler scheme for HPE.

\noindent
\textbf{Low/high resolution robustness.} Fig.~\ref{fig:robustness} visualizes the comparison results. Benefiting from low quantization error, SimCC-based methods can easily outperform heatmap-based counterparts in various input sizes (\emph{i.e.}, 64$\times$64, 128$\times$128, 256$\times$192 and 384$\times$288), demonstrating clear gains especially in low input resolutions. More specific quantitative results are discussed in Section~\ref{exp}.

\noindent
\textbf{Speed.} SimCC makes methods like~\cite{heatmap:xiao2018simple} get rid of time-consuming deconvolution modules, which can speed up the inference. It's worth noting that after removing the upsampling layers, SimpleBaseline-Res50~\cite{heatmap:xiao2018simple} with SimCC reduces \textbf{57.3\%} GFLOPs, improves \textbf{23.5\%} speed, and gains \textbf{+0.4} AP over heatmap-based counterpart. More comparisons are presented in Section~\ref{exp}.

%%%%%%%%%%%%%%%%%%%%%%%%%%%%%%% experiments
\section{Experiments}
\label{exp}
In the following sections, we empirically investigate the effectiveness of the proposed SimCC for 2D human pose estimation. We conduct experiments on three benchmark datasets: COCO~\cite{coco}, CrowdPose~\cite{li2019crowdpose}, and MPII~\cite{mpii}.

%%%%%%%%%%%%%%%%%%%%%%%%%%%%%%% coco
\subsection{COCO Keypoint Detection}
As one of the largest and most challenging datasets for HPE, the COCO dataset~\cite{coco} contains more than
200,000 images and 250,000 person instances labeling with 17 keypoints (\emph{e.g.}, nose, left ear, etc.). The COCO dataset is divided into three parts: 57k images for the training set, 5k for val set and 20k for test-dev set. In this paper, we follow the data augmentation in \cite{heatmap:sun2019deep}.

\vspace{.1cm} \noindent\textbf{Evaluation metric.} The standard average precision (AP) is used as our evaluation metric on the COCO dataset, which is calculated based on Object Keypoint Similarity (OKS):
\begin{equation}
    OKS=\frac{\sum_i exp(-d_i^2/2s^2j_i^2)\sigma(v_i>0)}{\sum_i \sigma(v_i>0)},
\end{equation}
where $d_i$ is the Euclidean distance between the $i$-th predicted keypoint coordinate and its corresponding coordinate groundtruth, $j_i$ is a constant, $v_i$ is the visibilty flag, $\sigma$ denotes indicator function and $s$ is the object scale.

\begin{table}[!ht]
\centering
\caption{\textbf{Comparisons with 2D heatmap-based methods on COCO validation set}. The results are provided with the same detected human boxes for fair comparison. 2D heatmap-based approaches adopt extra post-processing for refinement following their original paper, \emph{i.e.}, DARK~\cite{heatmap:darkpose} for TokenPose~\cite{heatmap:li2021tokenpose}  and empirical shift for HRNet~\cite{heatmap:sun2019deep} as well as SimpleBaseline~\cite{heatmap:xiao2018simple}. SimCC brings significant gains in all input resolutions while omitting costly refinement post-processing. In particular, SimCC-based SimBa-R50~\cite{heatmap:xiao2018simple} achieves better results than 2D heatmap-based counterpart with over \textbf{55\%} FLOPs reduction.}
\label{tab:val}
\scalebox{0.9}{
\begin{tabular}{lcccccll}
\toprule
\textbf{Method} & \textbf{Scheme} & \textbf{Input size} & \#\textbf{Params} &\textbf{GFLOPs}&\textbf{Extra post.} & \textbf{AP} & \textbf{AR }\\ \midrule
% tokenpose 64
\multirow{10}{*}{TokenPose-S \cite{heatmap:li2021tokenpose}} 
 & Heatmap & 64$\times$64 &4.9M &1.4 & w/o & 35.9   & 47.0  \\
& Heatmap & 64$\times$64 &4.9M & 1.4 & DARK~\cite{heatmap:darkpose}  & 57.1 & 64.8 \\
 &\textbf{SimCC} & 64$\times$64 & 4.9M & 1.4 & \textbf{w/o} & \textbf{62.8}  & \textbf{70.1}
 % tokenpose 128
 \\ \cmidrule{2-8} 
 & Heatmap & 128$\times$128 & 5.2M & 1.6 & w/o & 57.6  & 64.9  \\
  & Heatmap & 128$\times$128 & 5.2M & 1.6 & DARK~\cite{heatmap:darkpose} &  65.4 & 71.6 \\
 &\textbf{ SimCC} & 128$\times$128 &5.1M &1.6 &  \textbf{w/o} & \textbf{70.4} & \textbf{76.4} 
 % tokenpose 256
 \\ \cmidrule{2-8} 
 & Heatmap & 256$\times$192 & 6.6M &2.2 & w/o & 69.9  & 75.8  \\
   & Heatmap & 256$\times$192 & 6.6M &2.2 & DARK~\cite{heatmap:darkpose} & 72.5 & 78.0 \\
 & \textbf{SimCC} & 256$\times$192 & 5.5M & 2.2 & \textbf{w/o} & \textbf{73.6}  & \textbf{78.9}  \\ \midrule
 
  % simba-res50 
\multirow{10}{*}{SimBa-R50 \cite{heatmap:xiao2018simple}} 
& Heatmap & 64$\times$64 & 34.0M & 0.7 & w/o & 25.8  & 36.0  \\ 
& Heatmap & 64$\times$64 & 34.0M & 0.7& shift  & 34.4 & 43.7   \\
 & \textbf{SimCC }& 64$\times$64 & 24.7M & 0.3 & \textbf{w/o} & \textbf{39.3}  & \textbf{48.4} \\ \cmidrule{2-8} 
   & Heatmap & 128$\times$128 & 34.0M & 3.0 & w/o & 55.4   & 63.3  \\
 & Heatmap & 128$\times$128 & 34.0M & 3.0 & shift  & 60.3  & 67.6 \\
 & \textbf{SimCC}& 128$\times$128 & 25.0M & 1.3 & \textbf{w/o} & \textbf{62.6}  & \textbf{69.5}  \\ \cmidrule{2-8} 
   & Heatmap & 256$\times$192 & 34.0M & 8.9 & w/o & 68.5  & 74.8  \\
 & Heatmap & 256$\times$192 & 34.0M & 8.9 & shift & 70.4 & 76.3 \\
 & \textbf{SimCC} & 256$\times$192 & 25.7M & 3.8 & \textbf{w/o} & \textbf{70.8}  & \textbf{76.8}  \\ \midrule

 % hr48 64
\multirow{10}{*}{HRNet-W48 \cite{heatmap:sun2019deep}} 
 & Heatmap & 64$\times$64 & 63.6M & 1.2 & w/o & 36.9  & 47.8  \\
& Heatmap & 64$\times$64 &  63.6M &  1.2 & shift & 48.5 & 57.8 \\
 & \textbf{SimCC} & 64$\times$64 & 63.7M &1.2 & \textbf{w/o}  & \textbf{59.7}  & \textbf{67.5} % hr 128
 \\ \cmidrule{2-8} 
 & Heatmap & 128$\times$128 &  63.6M & 4.9 & w/o & 63.3  & 70.5  \\
  & Heatmap & 128$\times$128 &  63.6M & 4.9 & shift & 68.9 & 75.3 \\
 & \textbf{SimCC} & 128$\times$128 & 64.1M&4.9 &\textbf{w/o}   & \textbf{72.0}  & \textbf{77.9} 
 % hr256
 \\ \cmidrule{2-8} 
  & Heatmap & 256$\times$192 &  63.6M & 14.6 & w/o  & 73.1  & 78.7  \\
  & Heatmap & 256$\times$192 &  63.6M & 14.6 & shift & 75.1 & 80.4 \\
 & \textbf{SimCC} & 256$\times$192 & 66.3M & 14.6 & \textbf{w/o} & \textbf{75.9}  & \textbf{81.2}
 % hr384
  \\ 
 \bottomrule
\end{tabular}
}
\end{table}

\vspace{.1cm} \noindent\textbf{Baselines.} There are many \emph{CNN-based} and recent \emph{Transformer-based} methods for HPE. To show the superiority of the proposed SimCC, we choose two state-of-the-art methods (\emph{i.e.},  SimpleBaseline \cite{heatmap:xiao2018simple} and HRNet \cite{heatmap:sun2019deep}) from the former and one (\emph{i.e.}, TokenPose \cite{heatmap:li2021tokenpose}) from the latter as our baselines.

\vspace{.1cm} \noindent\textbf{Implementation details.} For the selected baselines, we simply follow the original settings in their papers. Specifically, for SimpleBaseline \cite{heatmap:xiao2018simple}, the base learning rate is set as $1e-3$, and is dropped to $1e-4$ and $1e-5$ at the 90-th and 120-th epochs respectively. For HRNet \cite{heatmap:sun2019deep}, the base learning rate is set as $1e-3$, and decreased to $1e-4$ and $1e-5$ at the 170-th and 200-th epochs. The total training processes are terminated within 140 and 210 epochs respectively for SimpleBaseline \cite{heatmap:xiao2018simple} and HRNet \cite{heatmap:sun2019deep}. Note that the training process of TokenPose-S follows \cite{heatmap:sun2019deep}. In this paper, we use the two-stage~\cite{heatmap:sun2019deep,heatmap:xiao2018simple,heatmap:chen2018cascaded,mppe} top-down human pose estimation pipeline: the person instances are firstly detected and then the keypoints are estimated. We adopt a popular person detector with $56.4\%$ AP provided by~\cite{heatmap:xiao2018simple} for COCO validation set. In addition, label smoothing is adopted in model training, which is commonly used in the task of classification for better generalization (equal smoothing sets the coefficient as 0.1 by default, following~\cite{labelSmoothing}). Experiments are conducted in 4 NVIDIA Tesla V100 GPUs.

\begin{table}[!ht]\footnotesize
  \centering
  \caption{\textbf{Comparisons on the COCO test-dev set}. `Trans.' represents Transformer~\cite{transformer} for short. ``$\dagger$" indicates that the Gaussian label smoothing is adopted. The proposed SimCC achieves state-of-the-art results, demonstrating clear performance improvements compared to 2D heatmap-based counterparts.}
  \label{tab:testdev}
  \scalebox{0.85}{
\begin{tabular}{llcclccccc}\toprule
\multicolumn{1}{l}{\textbf{Method}} &\textbf{Encoder} & \textbf{Input size} & \multicolumn{1}{c}{\textbf{GFLOPs}} & \multicolumn{1}{c}{\textbf{AP}} & \multicolumn{1}{c}{\textbf{AP$^{50}$}} & \multicolumn{1}{c}{\textbf{AP$^{75}$}} & \multicolumn{1}{c}{\textbf{AP$^{M}$}} & \multicolumn{1}{c}{\textbf{AP$^{L}$}} & \multicolumn{1}{c}{\textbf{AR}} \\ \midrule
\multicolumn{9}{c}{\textbf{2D Heatmap-based}} \\ \midrule
\multicolumn{1}{l}{Mask-RCNN~\cite{he2017mask}}& ResNet-50-FPN & - & \multicolumn{1}{c}{-} &  63.1  &  87.3  &  68.7  &  57.8  &  71.4  &  -  \\
\multicolumn{1}{l}{CMU-Pose~\cite{heatmap:cao2019openpose}}& VGG-19~\cite{VGG} & - & \multicolumn{1}{c}{-} &  64.2  &  86.2  &  70.1  &  61.0  &  68.8  &  -  \\
\multicolumn{1}{l}{G-RMI~\cite{mppe}}& ResNet-101~\cite{he2016deep} & 353$\times$257 & \multicolumn{1}{c}{-} &  64.9  &  85.5  &  71.3  &  62.3  &  70.0  &  69.7  \\
\multicolumn{1}{l}{AE~\cite{ae:newell2017associative}}& Hourglass~\cite{heatmap:newell2016stacked} & 512$\times$512 & \multicolumn{1}{c}{-}& 65.5 &86.8 &72.3 &60.6 &72.6 &70.2   \\
MultiPoseNet~\cite{kocabas2018multiposenet}&-&480$\times$480&-&69.6& 86.3& 76.6 &65.0& 76.3& 73.5\\
RMPE~\cite{heatmap:fang2017rmpe} & PyraNet~\cite{yang2017learning} &320$\times$256 & 26.7 &72.3 &89.2 &79.1 &68.0 &78.6 &-\\
\multicolumn{1}{l}{CPN~\cite{heatmap:chen2018cascaded}} & ResNet-Inception & 384$\times$288 & \multicolumn{1}{c}{29.2} &  72.1  &  91.4  &  80.0  &  68.7  &  77.2  &  78.5  \\
\multicolumn{1}{l}{CFN~\cite{CFN}} & - & - & \multicolumn{1}{c}{-} &  72.6  &  86.1  &  69.7  &  78.3  &  64.1  &  -  \\
\multicolumn{1}{l}{SimBa~\cite{heatmap:xiao2018simple}}& ResNet-152 & 384$\times$288 & \multicolumn{1}{c}{ 35.6} &  73.7  &  91.9  &  81.1  &  70.3  &  80.0  &  79.0  \\
\multicolumn{1}{l}{TransPose-H~\cite{heatmap:yang2020transpose}} &HRNet-W48+Trans.& 256$\times$192 & \multicolumn{1}{c}{ 21.8} & 75.0 & 92.2 & 82.3 & 71.3& 81.1&80.1 \\
 \multicolumn{1}{l}{HRNet-W32~\cite{heatmap:sun2019deep}}& HRNet-W32 & 384$\times$288 & \multicolumn{1}{c}{ 16.0} &  74.9  &  92.5  &  82.8  &  71.3  &  80.9  &  80.1  \\
\multicolumn{1}{l}{SimBa~\cite{heatmap:xiao2018simple}}& ResNet-50 & 384$\times $288 & \multicolumn{1}{c}{ 20.0} &   71.5 &  91.1  & 78.7 & 67.8  &  78.0 & 76.9   \\
\multicolumn{1}{l}{HRNet-W48~\cite{heatmap:sun2019deep}} &HRNet-W48& 256$\times$192 & \multicolumn{1}{c}{ 14.6} &  74.2  &  92.4  &  82.4  &  70.9  &  79.7  &  79.5  \\
\multicolumn{1}{l}{HRNet-W48~\cite{heatmap:sun2019deep}} &HRNet-W48& 384$\times$288 & \multicolumn{1}{c}{ 32.9} &  75.5  &  92.5  &  83.3  &  71.9  &  81.5  &  80.5\\
 \midrule
 
\multicolumn{9}{c}{\textbf{Regression-based}} \\ \midrule
\multicolumn{1}{l}{SPM~\cite{nie2019single}}& Hourglass~\cite{heatmap:newell2016stacked} &-& \multicolumn{1}{c}{-} & 66.9 & 88.5 & 72.9 & 62.6 & 73.1 & - \\
\multicolumn{1}{l}{DeepPose~\cite{toshev2014deeppose}}& ResNet-101 &256$\times$192 & \multicolumn{1}{c}{7.7} & 57.4 & 86.5 & 64.2 & 55.0 & 62.8 & - \\
\multicolumn{1}{l}{DeepPose~\cite{toshev2014deeppose}}& ResNet-152 &256$\times$192& \multicolumn{1}{c}{11.3} & 59.3 & 87.6 & 66.7 & 56.8 & 64.9 & - \\
\multicolumn{1}{l}{CenterNet~\cite{zhou2019objects}}& Hourglass~\cite{heatmap:newell2016stacked} &-& - & 63.0 &86.8& 69.6& 58.9 &70.4 & - \\
\multicolumn{1}{l}{DirectPose~\cite{tian2019directpose}}& ResNet-50 &-& \multicolumn{1}{c}{-} & 62.2 & 86.4 & 68.2 & 56.7 & 69.8 & - \\
\multicolumn{1}{l}{PointSetNet~\cite{pointset:wei2020point}}&HRNet-W48 &-& \multicolumn{1}{c}{-} & 68.7& 89.9& 76.3 &64.8 &75.3
 & - \\
 \multicolumn{1}{l}{Integral Pose~\cite{sun2018integral}}& ResNet-101 &256$\times$256& \multicolumn{1}{c}{11.0} & 67.8 & 88.2 & 74.8 & 63.9 & 74.0 & - \\
\multicolumn{1}{l}{TFPose~\cite{regression:mao2021tfpose}}& ResNet-50+Trans. &384$\times$288& \multicolumn{1}{c}{20.4} & 72.2 & 90.9 & 80.1 & 69.1 & 78.8 & - \\
\multicolumn{1}{l}{PRTR~\cite{li2021PRTR}}& HRNet-W48+Trans. &-& \multicolumn{1}{c}{-} & 64.9 & 87.0 & 71.7 & 60.2 & 72.5 & 74.1 \\
\multicolumn{1}{l}{PRTR~\cite{li2021PRTR}}& HRNet-W32+Trans. &384$\times$288& \multicolumn{1}{c}{21.6} & 71.7 & 90.6 & 79.6 & 67.6 & 78.4 & 78.8 \\
\multicolumn{1}{l}{PRTR~\cite{li2021PRTR}}& HRNet-W32+Trans. &512$\times$384& \multicolumn{1}{c}{37.8} & 72.1 & 90.4 & 79.6 & 68.1 & 79.0 & 79.4 \\ 
\multicolumn{1}{l}{RLE~\cite{rle}}& HRNet-W48 &-& \multicolumn{1}{c}{-} & 75.7 & 92.3 & 82.9 & 72.3 & 81.3 & - \\ 
 \midrule
 
\multicolumn{9}{c}{\textbf{SimCC-based}} \\ \midrule

\multicolumn{1}{l}{SimBa (SimCC$\dagger$)}& ResNet-50 &384$\times$288& \multicolumn{1}{c}{20.2} & 72.7 & 91.2 & 80.1 & 69.2 & 79.0 & 78.0 \\
\multicolumn{1}{l}{HRNet (SimCC$\dagger$)}& HRNet-W48 &256$\times$192& \multicolumn{1}{c}{14.6} & 75.4 & 92.4 & 82.7 & 71.9 & 81.3 & 80.5 \\
\multicolumn{1}{l}{HRNet (SimCC$\dagger$)}& HRNet-W48 &384$\times$288& \multicolumn{1}{c}{32.9} & \textbf{76.0}  & \textbf{92.4} & \textbf{83.5} & \textbf{72.5} & \textbf{81.9} & \textbf{81.1}\\
\bottomrule
\end{tabular}}
\end{table}

\noindent\textbf{Results on the COCO val set.} Extensive experiments are conducted on the COCO2017 validation set for comparing 2D heatmap-based and SimCC-based methods across various input resolutions (\emph{i.e.}, 64$\times$64, 128$\times$128, 256$\times$192, and 384$\times$288). Note that the evaluation and network training are under the same input size. Some top-performed CNN-based and Transformer-based methods are chosen as our baselines. Results presented in Table~\ref{tab:val} demonstrate that SimCC-based methods show consistent performance superiority over heatmap-based counterparts, especially in low-resolution input cases. For example, SimCC-based HRNet-W48~\cite{heatmap:sun2019deep} outperforms heatmap-based counterpart by \textbf{+0.8} AP at the input size of 256$\times$192. And under low input resolution as 64$\times$64, our SimCC shows much larger performance gain, \emph{i.e.}, \textbf{+11.2} AP on the COCO val dataset.

According to the results presented in Table~\ref{tab:val}, we can further draw the following conclusions: 1) Heatmap-based approaches rely seriously on post-processing for refinement, which brings extra computational cost and complicates the whole process. For example, TokenPose-S dramatically drops \textbf{21.2} AP at the input size of $64\times64$ if without the DARK~\cite{heatmap:darkpose} post-processing; 2) Our proposed SimCC works well without any refinement post-processing, leading to a more simple and efficient scheme compared to heatmap-based methods. For instance, our SimCC-based HRNet-W48 w/o extra post-processing outperforms heatmap-based counterpart (empirical shift is used) by 0.8 AP at the input size of $256\times192$.

\vspace{.1cm} \noindent\textbf{Results on the COCO test-dev set.} We conduct comparisons on COCO test-dev set and present the results in Table~\ref{tab:testdev}. SimCC-based HRNet-W48 and SimpleBaseline-Res50 surpass heatmap-based counterparts by \textbf{+0.5} and \textbf{+1.2} AP respectively, at the input size of 384$\times$288.

\vspace{.1cm} \noindent\textbf{Inference speed.} We discuss the impact of our proposed SimCC to the inference speed for SimpleBaseline~\cite{heatmap:xiao2018simple}, TokenPose-S~\cite{heatmap:li2021tokenpose} and HRNet-W$48$~\cite{heatmap:sun2019deep}. The `inference speed' here refers to the average time consuming of model feedforward (we compute 300 samples with batchsize=1). We adopt FPS to quantitatively illustrate the inference latency. The CPU implementation results are presented with the same machine: Intel(R) Xeon(R) Gold 6130 CPU @ 2.10GHz.

1) \emph{SimpleBaseline} Adopting SimCC allows one to remove the costly deconvolution layers of SimpleBaseline. We conduct experiments via SimpleBaseline-Res50~\cite{heatmap:xiao2018simple} on COCO val set with input size of $256\times192$. SimCC-based method w/o deconvolution modules can surpass 2D heatmap-based counterpart by \textbf{+0.4} AP (70.8 vs. 70.4) with \textbf{23.5\%} faster speed (21 vs. 17 FPS). More specific ablation study of deconvolution modules is conducted in Section~\ref{sec:ablation}.

2) \emph{TokenPose}\&\emph{HRNet} Due to that SimpleBaseline~\cite{heatmap:xiao2018simple} uses an encoder-decoder architecture, we can replace its decoder part (deconvolutions) with classifier heads of SimCC. But for HRNet~\cite{heatmap:sun2019deep} and TokenPose~\cite{heatmap:li2021tokenpose}, they have no extra independent modules as the decoder. To apply SimCC to them, we directly append the classifier heads to the original HRNet and replace the MLP head of TokenPose with ours, respectively. These are minor changes to the original architectures, thus only bringing little computation overhead for HRNet~\cite{heatmap:sun2019deep} and even reducing the model parameters for TokenPose~\cite{heatmap:li2021tokenpose}, as shown in Table~\ref{tab:val}. Hence, SimCC only has a slight impact on the inference latency for HRNet and TokenPose. For instance, the FPS of HRNet-W48 using heatmap or SimCC is almost the same (4.5/4.8) at the input size of 256$\times$192.    
\begin{table}[t]
  \centering
   \caption{\textbf{Comparisons with the 1D heatmap regression~\cite{yin2020attentive} and 2D heatmap-based methods}. Results achieved by SimBa-R50~\cite{heatmap:xiao2018simple} via different schemes (2D/1D heatmap, SimCC) on COCO val set with input size of $256\times256$. SimCC performs better than 2D/1D heatmap-based methods and requires \textbf{only 41.7\%/34.7\%} computation cost.}
   \label{tab:comparisons1dheatmap}
%   \resizebox{1\columnwidth}{!}{
\scalebox{0.9}{
\begin{tabular}{lccccc}
\toprule
\textbf{Scheme}  & \textbf{Deconvolation} & \textbf{\#Params} & \textbf{GFLOPs}  & \textbf{Extra post.} &\textbf{AP} \\ \midrule
2D Heatmap & 3  & 34.0M & 12.0 & shift &70.4 \\
2D Heatmap & 3  & 34.0M & 12.0 & w/o & 68.8 \\
1D Heatmap~\cite{yin2020attentive}  & 5  & 39.0M & 14.4  & w/o & 68.5  \\ 
\textbf{SimCC} & \textbf{0}  & \textbf{26.3M}  & \textbf{5.0}  & w/o &\textbf{70.4}  \\ 
\bottomrule
\end{tabular}

}
\end{table}

\noindent\textbf{Is 1D heatmap regression a promising solution for HPE?} We also study the performance of expanding the 1D heatmap~\cite{yin2020attentive} into the field of HPE, which is initially designed for facial landmark. Table~\ref{tab:comparisons1dheatmap} shows that the 1D heatmap regression~\cite{yin2020attentive} increases the model parameters and computational cost yet performs even worse than 2D heatmap-based counterpart. The potential reason might be that the core challenges of facial landmark and HPE are different: facial landmark possesses rigid deformation while human body joints have much higher degrees of freedom. Since the co-attention module as well as decoupling layers in~\cite{yin2020attentive} are only empirically verified to be effective for the task of facial landmark and their generalization to other fields (\emph{e.g.}, HPE) remains unclear.

%%%%%%%%%%%%%%%%%%%%%%%%%%%%%%% ablation study
\subsection{Ablation Study}
\label{sec:ablation}
\begin{figure}[t]
\centering
\subfigure[128$\times$128 input size.]{
\includegraphics[width=2.2in]{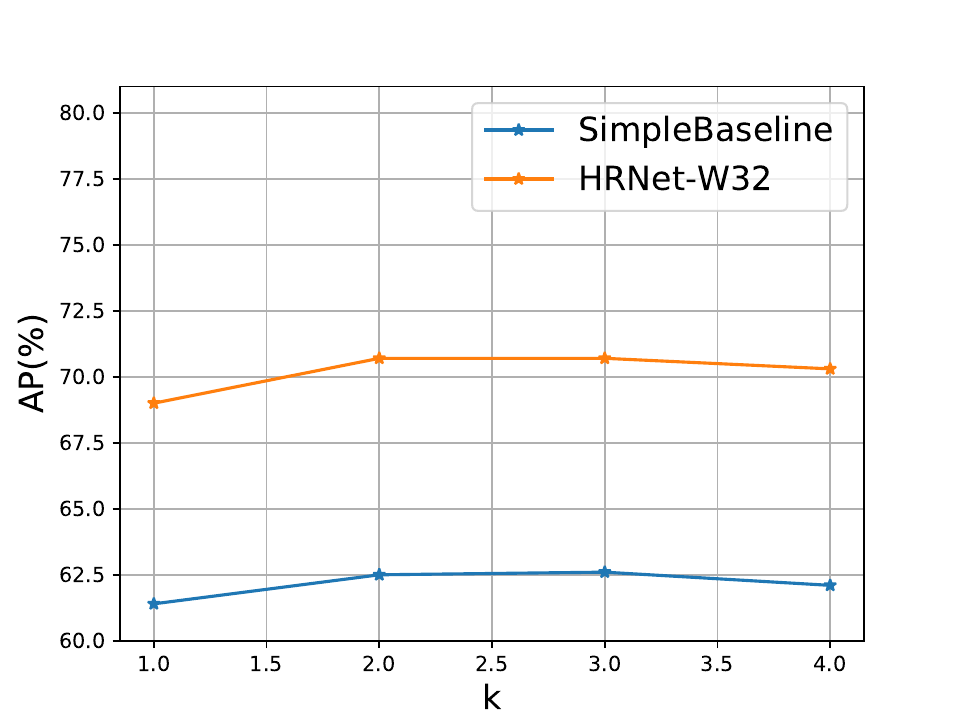}
}
\quad
\subfigure[256$\times$192 input size.]{
\includegraphics[width=2.2in]{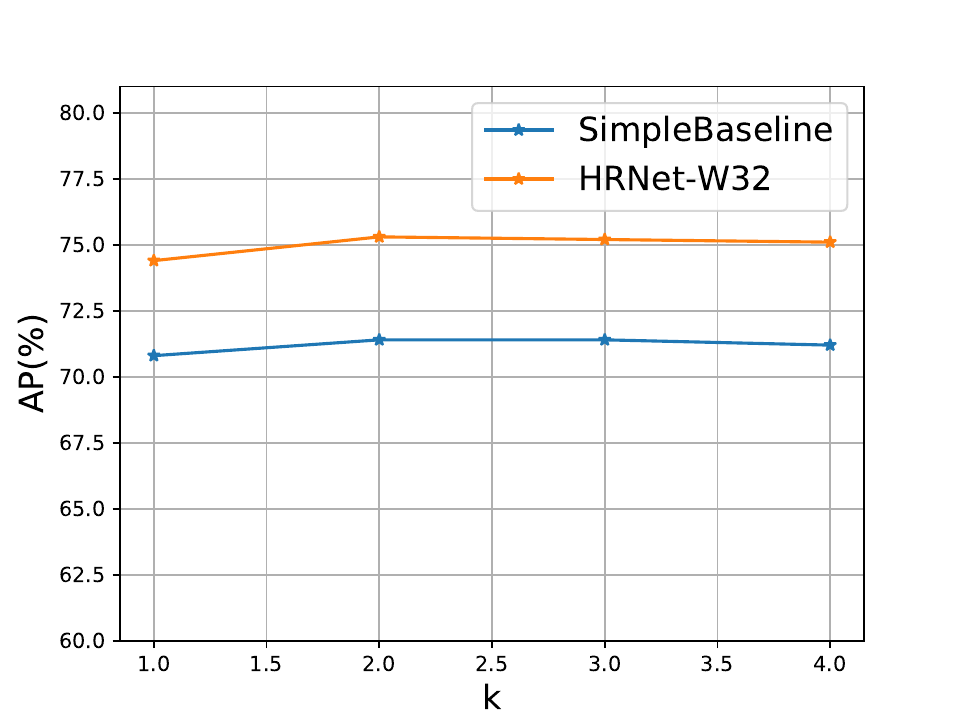}
}
\caption{\textbf{Ablation study of splitting factor $k$ value on the COCO validation set}. SimpleBaseline~\cite{heatmap:xiao2018simple} uses ResNet-50 as backbone. $k$ controls how many bins per pixel in SimCC. Model performance increases first and then drops as $k$ becomes larger.}
\label{fig:ablation}
\end{figure}

\vspace{.1cm} \noindent\textbf{Splitting factor $k$.}
The splitting factor $k$ controls the how many bins per pixel in SimCC. Specifically, the larger $k$ is, the smaller the quantization error of SimCC is. Nevertheless, model training becomes more difficult when $k$ increases. Hence, there is a trade-off between the quantization error and the model performance. We test $k\in\{1,2,3,4\}$ based on SimpleBaseline~\cite{heatmap:xiao2018simple} and HRNet~\cite{heatmap:sun2019deep} under various input resolutions. As shown in Fig.~\ref{fig:ablation}, model performance tends to increase first and then decrease as $k$ grows. For HRNet-W32~\cite{heatmap:sun2019deep}, the recommended settings are $k=2$ for both 128$\times$128 and 256$\times$192 input size. For SimBa-Res50~\cite{heatmap:xiao2018simple}, the recommended settings are $k=3$ and $k=2$ for 128$\times$128 and 256$\times$192 input size, respectively.

\begin{table}[!t]
  \centering
   \caption{\textbf{Ablation study of upsampling modules}. Results achieved by SimBa-R50~\cite{heatmap:xiao2018simple} on COCO val set. ``Heatmap" represents 2D heatmap-based methods for short. Employing deconvolution improves SimCC-based methods yet the gains are slight. Even without any deconvolution layers, SimCC-based approaches surpass 2D heatmap-based counterparts, significantly reducing over 55\% FLOPs.}
   \label{tab:up}
\scalebox{1.0}{
\tabcolsep0.05in\centering
\begin{tabular}{cccccc}
\toprule
\textbf{Scheme} & \textbf{Input size} & \textbf{Deconvolution} & \textbf{\#Params} & \textbf{GFLOPs} & \textbf{AP} \\ \midrule
Heatmap & 64$\times$64 & \checkmark & 34.0M & 0.7 & 34.4 \\
\textbf{SimCC}  & 64$\times$64 & \checkmark & 34.1M & 0.7  & \textbf{40.8}  \\ 

\textbf{SimCC} & 64$\times$64 & \ding{55} & \textbf{24.7M}  & \textbf{0.3}  & 39.3  \\ 
 \hline 
 
Heatmap & 128$\times$128 & \checkmark & 34.0M & 3.0 & 60.3 \\
\textbf{SimCC} & 128$\times$128 & \checkmark & 34.8M & 3.0  & \textbf{62.6} \\ 

\textbf{SimCC} & 128$\times$128 & \ding{55} & \textbf{25.0M}   & \textbf{1.3}   & \textbf{62.6}   \\ 
 \hline
 
Heatmap & 256$\times$192 & \checkmark & 34.0M & 8.9 & 70.4 \\
\textbf{SimCC} & 256$\times$192 & \checkmark & 36.8M & 9.0  & \textbf{71.4} \\ 
 
\textbf{SimCC} & 256$\times$192 & \ding{55} & \textbf{25.7M}  & \textbf{3.8}  & 70.8 \\
\bottomrule
\end{tabular}
}
\end{table}
\begin{table}[!t]
  \centering
    \caption{\textbf{Ablation study of label smoothing}. Results are achieved based on SimpleBaseline-Res50~\cite{heatmap:xiao2018simple} with the input size of 384$\times$288 on COCO2017 val dataset. Employing label smoothing significantly improves the performance by 2.1 AP.}
   \label{tab:dis}
\scalebox{1.0}{
\tabcolsep0.07in\centering
\begin{tabular}{ccccccc}
\toprule
\textbf{Label smoothing} & \textbf{AP} & \textbf{AP$^{50}$} & \textbf{AP$^{75}$}& \textbf{AP$^{M}$} & \textbf{AP$^L$} & \textbf{AR} \\ \midrule
w/o & 71.3 & 88.8 & 78.2 & 67.8 & 78.2 & 77.3 \\
equal  & 73.0 & 89.3 & 79.7 & 69.5 & 79.9 & 78.6 \\
Gaussian  & 73.4 & 89.2 & 80.0 & 69.7 & 80.6 & 78.8 \\ 
Laplace  & 73.0 & 89.3 & 79.7 & 69.3 & 80.3 & 78.4 \\ 
\bottomrule
\end{tabular}
}
\end{table}

\vspace{.1cm} \noindent\textbf{Upsampling modules.} Upsampling modules are usually computational costly and substantially slow down the network's inference speed, however, indispensable for heatmap-based methods. Hence, it's of practical significance to explore if applying SimCC can reduce the dependence of upsampling modules in HPE. Notice that the upsampling modules\footnote{The upsampling modules used in SimpleBaseline~\cite{heatmap:xiao2018simple} recover the feature map resolution from 1/32$\times$ to 1/4$\times$ input size, consisting of three deconvolution layers.} adopted in SimpleBaseline~\cite{heatmap:xiao2018simple} is independent to the backbone and thus can be easily removed. Therefore we conduct ablation study of SimCC w/ and w/o deconvolution modules based on SimpleBaseline~\cite{heatmap:xiao2018simple}. Table~\ref{tab:up} show the results on the COCO2017 val dataset. It can be observed that compared to heatmap, SimCC allows one to remove the costly deconvolution layers of SimpleBaseline, resulting in consistent computational cost reduction across various input resolutions. For example, SimCC-based SimpleBaseline-Res50 w/o upsampling modules still outperform heatmap-based counterpart w/ upsampling modules by \textbf{+0.4} AP, with \textbf{57.3\%} fewer GFLOPs at the input size of 256$\times$192.

\noindent\textbf{Label smoothing.} Label smoothing~\cite{labelSmoothing} is a commonly used strategy to improve generalization for the task of classification. To investigate its effect on our proposed method, we train SimpleBaseline-Res50~\cite{heatmap:xiao2018simple} based on SimCC with various label smoothing strategies: \{\emph{w/o}, \emph{equal}, \emph{Gaussian}, \emph{Laplace}\}. Table~\ref{tab:dis} demonstrates that label smoothing strategy does make a difference. Therefore, a promising way to further improve SimCC may be replacing the heuristic label smoothing strategy in a self-adaptive way. Further discussion is out the scope of this paper and we regard it as future work.
    
\begin{table}[!ht]
  \centering
  \caption{\textbf{Comparisons with 2D heatmap-based methods on CrowdPose test dataset}. Results are achieved by HRNet-W32~\cite{heatmap:sun2019deep} and ``Heatmap" denotes 2D heatmap as an abbreviation. SimCC-based HRNet-W32 demonstrates consistent improvements compared to 2D heatmap-based methods.}
  \label{tab:crowd}
 \scalebox{1.0}{
    \tabcolsep0.07in\centering
\begin{tabular}{cccccccc}
\toprule
\textbf{Scheme} & \textbf{Input size} & \textbf{AP} & \textbf{AP$^{50}$} & \textbf{AP$^{75}$} & \textbf{AP$^E$} & \textbf{AP$^M$} & \textbf{AP$^H$} \\ \midrule
 Heatmap & 64$\times$64 & 42.4 & 69.6 & 45.5 & 51.2  & 43.1& 31.8 \\
 SimCC  & 64$\times$64 & \textbf{46.5} & \textbf{70.9} & \textbf{50.0} & \textbf{56.0} & \textbf{47.5} & \textbf{34.7} \\ \hline
 Heatmap & 256$\times$192 & 66.4 & 81.1 & 71.5 & 74.0  & 67.4& 55.6 \\
 SimCC  & 256$\times$192 & \textbf{66.7} & \textbf{82.1} & \textbf{72.0} & \textbf{74.1} & \textbf{67.8} & \textbf{56.2} \\
 \bottomrule
\end{tabular}
}
\end{table}

%%%%%%%%%%%%%%%%%%%%%%%%%%%%%%% crowedpose
\subsection{CrowdPose}

One may concern about the performance of SimCC in dense pose scenes. Thus we further illustrate the effectiveness of the proposed SimCC on the CrowdPose~\cite{li2019crowdpose} dataset, which contains much more crowded scenes than the COCO keypoint dataset. There are 20K images and 80K person instances in the CrowdPose. The training, validation and testing subset consist of about 10K, 2K, and 8K images respectively. Similar evaluation metric to that of COCO~\cite{coco} is adopted here, with extra $AP^{E}$ (AP scores on relatively easier samples) and $AP^{H}$ (AP scores on harder samples). We follow the original paper~\cite{li2019crowdpose} to adopt YoloV3~\cite{redmon2018yolov3} as the human detector, and batch size is set as 64. We conduct comparison experiments on the CrowdPose test dataset, at the input size of 64$\times$64 and 256$\times$192 respectively. The results in Table~\ref{tab:crowd} show that SimCC-based methods outperform heatmap-based counterparts.

%%%%%%%%%%%%%%%%%%%%%%%%%%%%%%% mpii
\subsection{MPII Human Pose Estimation}
The MPII Human Pose dataset~\cite{mpii} contains 40K person samples with 16 joints labels. We point out that the data augmentation used on the MPII dataset is the same as that on COCO dataset.

\begin{table}[!ht]
  \centering
    \caption{\textbf{Comparisons with 2D heatmap-based methods on the MPII validation set}. Experiments are conducted based on HRNet-W32~\cite{heatmap:sun2019deep} and  ``Heatmap" means 2D heatmap for short. ``$\dagger$" denotes that Gaussian label smoothing is utilized. Under stricter metric PCKh@0.1, SimCC shows clear gains across different input sizes.}
  \label{tab:mpii-val}
  \scalebox{1.0}{
      \tabcolsep0.04in\centering
\begin{tabular}{cccccccccc}
\toprule
\textbf{Scheme} & \textbf{Input size} & \textbf{Hea} & \textbf{Sho} & \textbf{Elb} & \textbf{Wri} & \textbf{Hip} & \textbf{Kne} & \textbf{Ank} & \textbf{Mean}  \\ \midrule
\multicolumn{10}{c}{\textbf{PCKh@0.5}} \\ \midrule
Heatmap & 64$\times$64 & 89.7 & 86.6 & 75.1 & 65.7 & 77.2 & 69.2 & 63.6 & 76.4   \\
SimCC & 64$\times$64& \textbf{93.5} & \textbf{89.5} & \textbf{77.5} & \textbf{67.6} & \textbf{79.8} & \textbf{71.5} & \textbf{65.0} & \textbf{78.7}  \\ \hline
Heatmap & 256$\times$256  & 97.1 & 95.9 & 90.3 & \textbf{86.4} & 89.1 & \textbf{87.1} & \textbf{83.3} & \textbf{90.3} \\
SimCC & 256$\times$256 & 96.8 & 95.9 & 90.0 & 85.0 & 89.1 & 85.4 & 81.3 & 89.6  \\
SimCC$\dagger$ & 256$\times$256 & \textbf{97.2} & \textbf{96.0} & \textbf{90.4} & 85.6 & \textbf{89.5} & 85.8 & 81.8 & 90.0  \\ \midrule

\multicolumn{10}{c}{\textbf{PCKh@0.1}} \\ \midrule
Heatmap & 64$\times$64 & 12.9 & 11.7 & 9.7 & 7.1 & 7.2 & 7.2 & 6.6 & 9.2 \\
SimCC  & 64$\times$64 & \textbf{30.9} & \textbf{23.3} & \textbf{18.1} & \textbf{15.0} & \textbf{10.5} & \textbf{13.1} & \textbf{12.8} & \textbf{18.5} \\ \hline
Heatmap & 256$\times$256 & 44.5 & 37.3 & 37.5 & 36.9 & 15.1 & 25.9 & 27.2 & 33.1 \\
SimCC & 256$\times$256 & \textbf{50.1} & 41.0 & \textbf{45.3} & \textbf{42.4} & 16.6 & \textbf{29.7} & \textbf{30.3} & \textbf{37.8} \\ 
SimCC$\dagger$ & 256$\times$256 & 49.6 & \textbf{41.9} & 43.0 & 39.6 & \textbf{17.0} & 28.2 & 28.9 & 36.8 \\ 
\bottomrule
\end{tabular}
      }
\end{table}

\vspace{.1cm} \noindent\textbf{Results on the validatoin set.} We follow the evaluation procedure in HRNet~\cite{heatmap:sun2019deep}. The head-normalized probability of correct keypoint (PCKh)~\cite{mpii} score is used for model evaluation. The results are presented in Table~\ref{tab:mpii-val}. At the input size of 256$\times$256, SimCC-based methods achieve competitive performances under PCKh@0.5, and show clear gains under the stricter measurement PCKh@0.1.

\section{Limitation and Future Work}
\label{discussion}
SimCC introduced in this paper works under the setting of top-down human pose estimation. When it comes to bottom-up multi-person pose estimation, the presence of multiple people brings the identification ambiguity. Potential future work is to introduce extra embeddings in a similar way to AE~\cite{ae:newell2017associative}, in order to address the matching problem between candidate coordinate $x$ and $y$ values.

\section{Conclusion}
In this paper, we explore a simple yet promising coordinate representation (namely SimCC). It regards the keypoint localization task as two sub-tasks of classification for horizontal and vertical axes, representing the $x$- and $y$- coordinate of joint location into two independent 1D vectors. The experimental results empirically show that the 2D structure might not be a key ingredient for coordinate representation to sustain superior performance. The proposed SimCC shows advantages over heatmap-based representation at model performances. Moreover, it may also inspire new works on lightweight model design for HPE.

\clearpage
\bibliographystyle{splncs04}
\bibliography{egbib}
\end{document}